\documentclass[sigconf]{acmart}
\usepackage{subcaption}

\AtBeginDocument{%
  \providecommand\BibTeX{{%
    \normalfont B\kern-0.5em{\scshape i\kern-0.25em b}\kern-0.8em\TeX}}}

\setcopyright{acmlicensed}
\copyrightyear{2024}
\acmYear{2024}
\acmDOI{XXXXXXX.XXXXXXX}

\acmConference[COMPSCI 280 '24]{UC Berkeley EECS}{May '24}{Berkeley, CA}
\acmISBN{978-1-4503-XXXX-X/24/04}

\begin{document}

\title{Hologram: Realtime Holographic Overlays via LiDAR Augmented Reconstruction}

\author{Ekansh Agrawal}
\affiliation{%
  \institution{UC Berkeley EECS}
  \country{USA}}
\email{agrawalekansh@berkeley.edu}

\begin{abstract}
Guided by the hologram technology of the infamous Star Wars franchise, I present an application that creates real-time holographic overlays using LiDAR augmented 3D reconstruction. Prior attempts involve SLAM or NeRFs which either require highly calibrated scenes, incur steep computation costs, or fail to render dynamic scenes. I propose 3 high-fidelity reconstruction tools that can run on a portable device, such as a iPhone 14 Pro, which can allow for metric accurate facial reconstructions. My systems enable interactive and immersive holographic experiences that can be used for a wide range of applications, including augmented reality, telepresence, and entertainment.
\end{abstract}





\maketitle

\section{Motivation}

As a kid, an engineer, and a researcher, the world of Stars Wars has never stopped to pique my curiousity. Whether it's the fast-than-light intergalatic space travel or the iconic laser sword "lightsabers", the Star Wars universe has always been a source of inspiration for me. One of the most iconic technologies in the Star Wars universe is the hologram as shown in. Soldiers, politicians, and civilians alike all use portable devices to project 3D images of people or objects in real-time to communicate in real-time. This technology has been a staple in the Star Wars franchise and in this project, I aim to bring this technology to life. My goal is to create a system that can project real-time holographic overlays using LiDAR augmented 3D reconstruction. The idea of a hologram is not just a cool visual effect, but a powerful tool that can be used for a wide range of applications, including augmented reality, telepresence, and entertainment.

While there have been many notable hardware advancements in 3D reconstructions and holographic displays, there are still many challenges that need to be addressed before they can be proposed as an everyday consumer electronic \cite{lucente1994holographic} \cite{lucente1995optimization} \cite{peng2020deep} \cite{shi2021towards}. For the scope of this project, I focus solely on software as a vehicle of real-time reconstructions. I also further narrow the scope by mostly focusing on trying to reconstruct faces. Existing methods rely on highly calibrated environments, incur steep computation costs, or fail to render dynamic scenes \cite{zander2012computational}. In this project, I propose three high-fidelity reconstruction tools that can run on a portable device, such as an iPhone 14 Pro, which can allow for metric accurate facial reconstructions. These systems enable interactive and immersive holographic experiences that can be used for a wide range of applications, including augmented reality, telepresence, and entertainment.

\begin{figure}
    \centering
    \includegraphics[width=8cm]{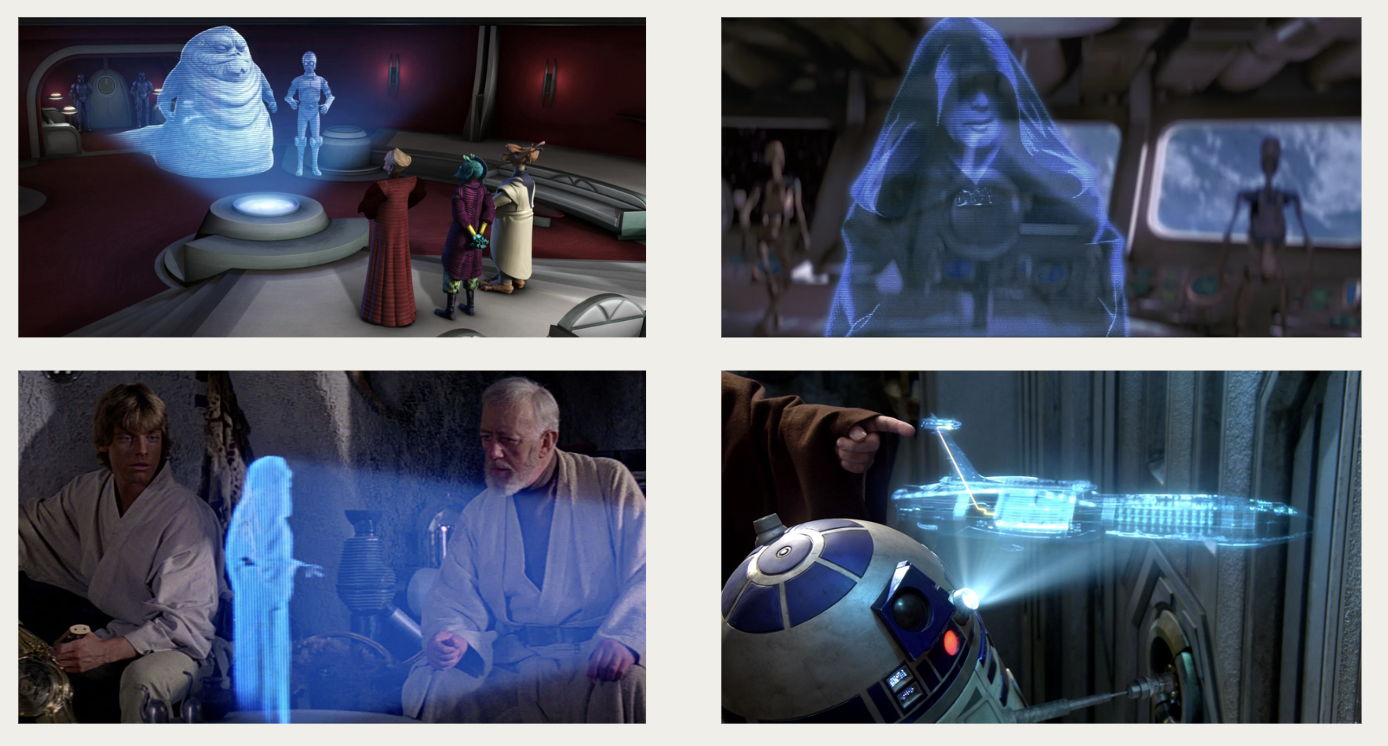}
    \caption{Hologram technology from the Star Wars universe}
    \label{fig:starwars}
\end{figure}

\section{Prior Works}

\subsection{Simultaneous Localization and Mapping}

With the introduction of the Kinect Sensor, low-resolution depth and visual sensing introduced novel approaches for real-time real-time dense 3D reconstruction \cite{kinectcv}. Initial research employs a volumetric truncated signed distance function (TSDF) to fuse depth data into a global model. KinectFusion achieves impressive results but is limited to relatively small-scale environments due to memory constraints \cite{kinectfusion}. ElasticFusion extends KinectFusion by introducing a more flexible and scalable map representation using surfels (surface elements) \cite{elasticfusion}. It can handle larger environments and loop closures, enabling robust and consistent reconstructions. Real-time 3D reconstruction system that combines depth and RGB data. It leverages the complementary strengths of geometric and photometric information, resulting in high-quality reconstructions with accurate geometry and detailed texture \cite{bundlefusion}. Research has allowed large-scale, real-time 3D reconstructions via multiple Kinect sensors by efficiently allocating and updating only the occupied regions in the volume \cite{voxelhashing}  \cite{staticfusion} \cite{infinitam}.

While these methods have made significant progress in real-time 3D reconstruction, they are limited to static environments and struggle to handle dynamic scenes in high resolution. This makes live reconstruction of a high detail object like a subject's face temperamental. They rely on geometric priors and are sensitive to noise and outliers in the input data. Furthermore, most of these methods rely on the Kinect Sensor which is not as portable as the project goal.

\subsection{Depth Estimation}

Depth estimation of a scene allows us to perform classical image transformations to scale coordinates from the image plane to the world plane. This is a crucial step in 3D reconstruction as it allows us to map the 2D image to a 3D point cloud quite effortlessly given the frame depth and calibration parameters. Traditional methods for depth estimation include stereo matching, structure from motion, and depth from de-focus. Stereo matching involves finding correspondences between two images taken from different viewpoints and computing a disparity map of some kind \cite{scharstein2003high} \cite{birchfield1999depth}. This techniques usually are contingent on a highly calibrated environment in order to use epipolar geometry to extrapolate between the 3D points and their projections onto the 2D images \cite{kyto2011method} \cite{marr1976cooperative}. More recent advances in stereo vision include the use of CNNs for disparity estimation \cite{zbontar2016stereo} \cite{kendall2017geometriccnn}. Significant progress has also been made in the realm of monocular depth estimation via deep learning approaches that learn to extract relevant features and depth cues from large-scale datasets of RGB images and corresponding depth maps \cite{eigen2014depth} \cite{liu2015learning}. Advancements like encoder-decoder architectures \cite{laina2016deeper} \cite{hu2019revisiting}, attention mechanisms and multi-task learning \cite{Chen2019aaai} \cite{wang2020dfnet}, transformers-based architectures \cite{ranftl2021vision} \cite{chen2022monodeptformers}, and even diffusion-based architecture \cite{saxena2023monocular} \cite{ke2023repurposing} have all attempted to reduce the disparity of a neural based approach for depth estimation.

Structure from motion techniques aim to reconstruct 3D structures from a sequence of 2D images captured from different viewpoints. Basic techniques involve extrapolating an essential matrix between the images and then using the epipolar geometry to triangulate the 3D points \cite{longuet1981computer} \cite{tomasi1992shape} \cite{pollefeys2008detailed}. More recent advances in structure from motion include the use of deep learning techniques to estimate the camera poses and 3D points \cite{ummenhofer2017demon} \cite{zhou2017unsupervised}. My major qualm with adopting these techniques is the need for a highly calibrated environment. These techniques also rely heavily on using more than one camera to capture the scene \cite{zhao2018panoramic}. I introduce engineering constraints later in this project which make it difficult to adopt this approach.

3D reconstructions via LiDAR involve processing and interpreting these point cloud data to create accurate 3D models or representations of objects, scenes, or environments. Through time-of-flight calculations, LiDAR technology can create efficient point cloud registration \cite{besl1992method} \cite{yang2016go} which can be used for surface reconstruction \cite{kazhdan2006poisson} \cite{oztireli2009topologyrepair}. Deep learning approaches \cite{huang2018recurrent} \cite{jiang2022lidar} are instrumental for large-scale scene reconstruction \cite{pan2013probuilding} \cite{oehler2011efficient}. The main problem with consumer grade LiDAR sensors are their lack of precision \cite{ma2022lidar}, and industrial grade sensors can cost tens of thousands of dollars making their usage cost prohibitive \cite{yang2018towards}.

\subsection{Neural Radiance Fields}

The advent of NeRFs introduced an approach to synthesizing novel views \cite{mildenhall2020nerf}, highlighting their abilities to capture and render complex scenes with high fidelity \cite{mueller2022instant} \cite{tancik2022barf}, including intricate geometric details, view-dependent effects (e.g., specularity, transparency), and challenging lighting conditions \cite{trevithick2021grf} \cite{yu2021pixelnerf}. This is achieved by the high-dimensional and continuous nature of the learned representation, which can encode rich scene properties without relying on explicit 3D geometry or texture mapping \cite{mildenhall2020nerf}. NeRFs have been extended to handle dynamic scenes by incorporating temporal information \cite{li2021neural} \cite{pumarola2021d}, enabling the reconstruction of moving objects and scenes. Further optimizations introduced high quality renders with sparse input views \cite{garbin2021efficient} \cite{liu2020neural} and close to real-time rendering speeds for individual scenes \cite{hedman2021snerg} \cite{rebain2022derf} \cite{sitzmann2020implicit} \cite{srinivasan2021nerv}. However, in order to achieve  high-fidelity facial reconstructions, NeRFs require a large number of input views and high-resolution images, which can be computationally expensive and time-consuming. This makes it difficult to achieve real-time performance on portable devices, like the iPhone 14 Pro, since I would need to render a new scene for every frame.

\section{Methodology}

\begin{figure*}
    \centering
    \includegraphics[width=0.9\textwidth]{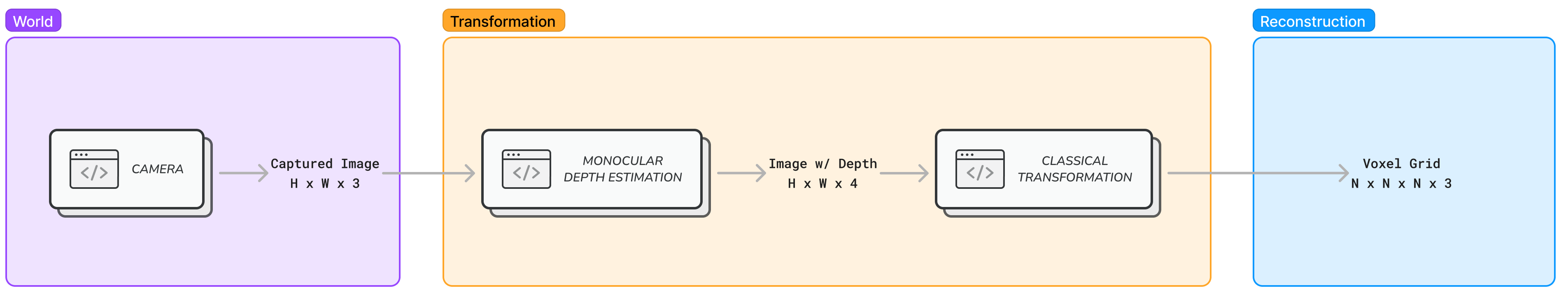}
    \caption{Attempt 1 utilizes monocular depth estimation alongside classical image projections in order to transfer pixels from their image representation into there voxel representation.}
    \label{fig:attempt1}
\end{figure*}

\begin{figure*}
    \centering
    \includegraphics[width=0.9\textwidth]{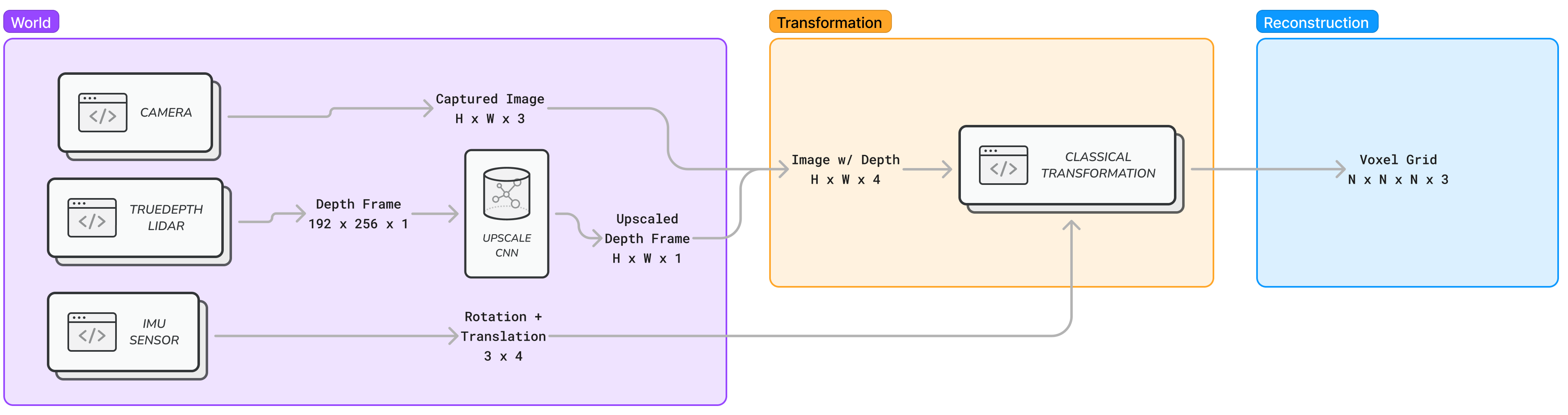}
    \caption{Attempt 2 utilizes the LiDAR and TrueDepth data streamed from the iPhone 14 Pro's front facing sensors. This data is then upscaled with a SRCNN model trained on facial depth maps. I fuse this data and then use classical projections using the phone's IMU sensor for camera poses.}
    \label{fig:attempt2}
\end{figure*}

\begin{figure*}
    \centering
    \includegraphics[width=0.9\textwidth]{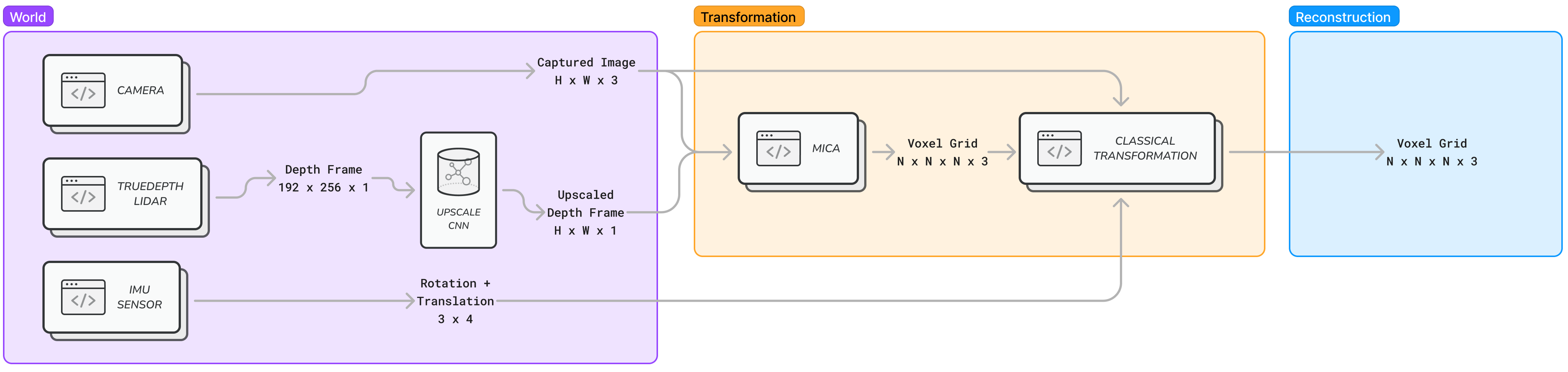}
    \caption{Attempt 3}
    \label{fig:attempt3}
\end{figure*}

I propose three approaches to generating real-time 3D facial reconstructions in this paper. While these individual techniques may not be novel in nature, the hybrid approach along with the constraint of the real-time reconstruction requires quite a bit of engineering to get them to work with high fidelity. I self-imposed the following engineering constraints on my methodology to ensure that the engineered solutions were novel and challenging to implement:

\begin{enumerate}
    \item \textbf{Monocular estimation}: Reconstruction cannot be performed with more than a single set of sensors. (IE. only one camera, one LiDAR sensor, etc..)
    \item \textbf{Portability}: If the clone troopers carried around tiny portable devices, I want to be able to do same. No backup RTX 4090s.
    \item \textbf{High fidelity}: The final rendered voxel grid cannot be insanely sparse. It should be extremely clear what's being rendered, and we should be able to tell exactly who's face is being reconstructed.
    \item \textbf{Instantaneous}: A viable solution must achieve at least 30 FPS during reconstruction and must occur online.
\end{enumerate}

\subsection{Attempt 1: Monocular Depth Estimation}

My initial approach, as shown in Figure \ref{fig:attempt1}, is quite simple in nature. Given an image frame, I want to use our classical projection formulas in order to transform image from the \textit{uv} coordinate system to the world coordinate system. Calibrated environments can utilize stereo imaging to leverage disparity maps to compute depth over a calibrated scene. I can replace this depth approach with a monocular-depth estimation model. I use Intel MiDaS as the core backbone of my depth estimation pipeline, consisting of a ViT backbone pre-trained off of ImageNet \cite{ranftl2020midas} \cite{birkl2023midas}. A single frame is passed into this model, and the output is a $H$ by $W$ matrix consisting of the relative depth of each pixel from the input image frame as $z_c$. I then extract the camera intrinsics from the webcam and apply Equation \ref{eq:1} to extract the camera points. I use identify transformations and apply Equation \ref{eq:2} to project the points into the world coordinate system.

\begin{equation} \label{eq:1}
    \begin{bmatrix}
    x_c \\
    y_c \\
    z_c
    \end{bmatrix}
    =
    \begin{bmatrix}
    f_x & 0 & c_x \\
    0 & f_y & c_y \\
    0 & 0 & 1
    \end{bmatrix}
    \begin{bmatrix}
    u \\
    v \\
    1
    \end{bmatrix}
    +
    \begin{bmatrix}
    0 \\
    0 \\
    z_c - 1
    \end{bmatrix}
\end{equation}

\begin{equation} \label{eq:2}
    \begin{bmatrix}
    x_w \\
    y_w \\
    z_w
    \end{bmatrix}
    = R
    \begin{bmatrix}
    x_c \\
    y_c \\
    z_c
    \end{bmatrix}
    + T
\end{equation}

As a means of comparison, I also used the Marigold diffusion model for depth estimation which utilizes a fine-tuned Stable Diffusion model \cite{ke2023repurposing}. I used 10 denoising steps and preserved the image resolution during model inferences.

\subsection{Attempt 2: LiDAR + TrueDepth}

\begin{figure}
    \centering
    \includegraphics[width=8cm]{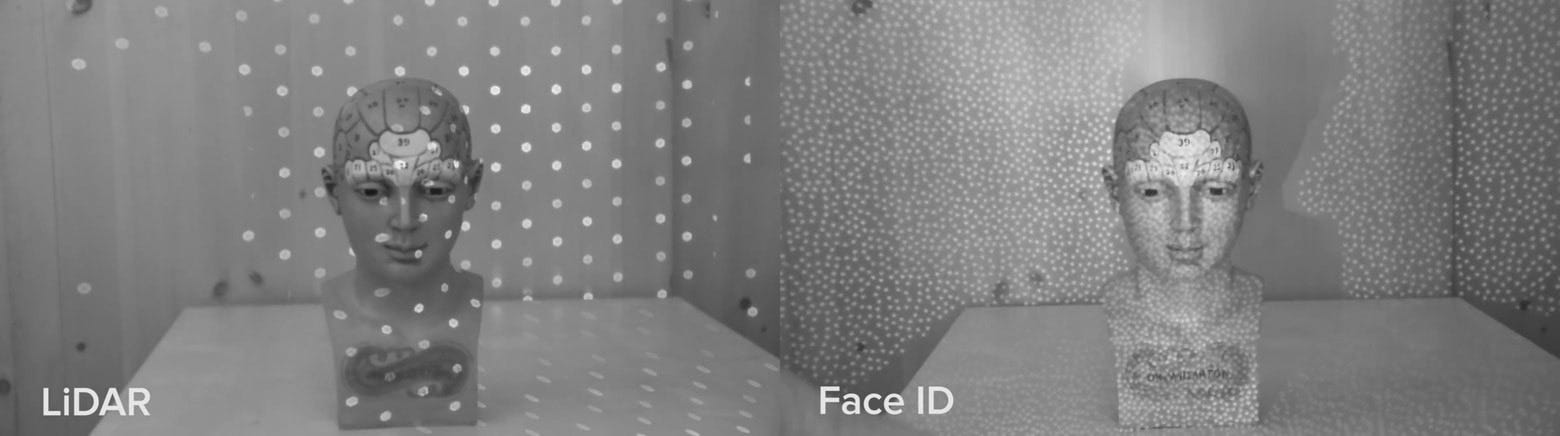}
    \caption{Projected light coordinates for the TrueDepth and LiDAR sensors shipped with the Apple iPhones.}
    \label{fig:truedepth}
\end{figure}

The release of Apple's iPhone 12 Pro introduced a consumer-grade LiDAR sensor and TrueDepth infrared depth sensor. The combination of these sensors project points onto the subject and record the time it takes for points to reflect light back into the sensors \cite{luetzenburg2021evaluation}. The LiDAR sensor works well with coarse objects while the TrueDepth sensors works well with differentiating distance between fine objects as shown in Figure \ref{fig:truedepth}.

\begin{figure}
     \centering
     \begin{subfigure}[t]{0.15\textwidth}
         \centering
         \includegraphics[width=\textwidth]{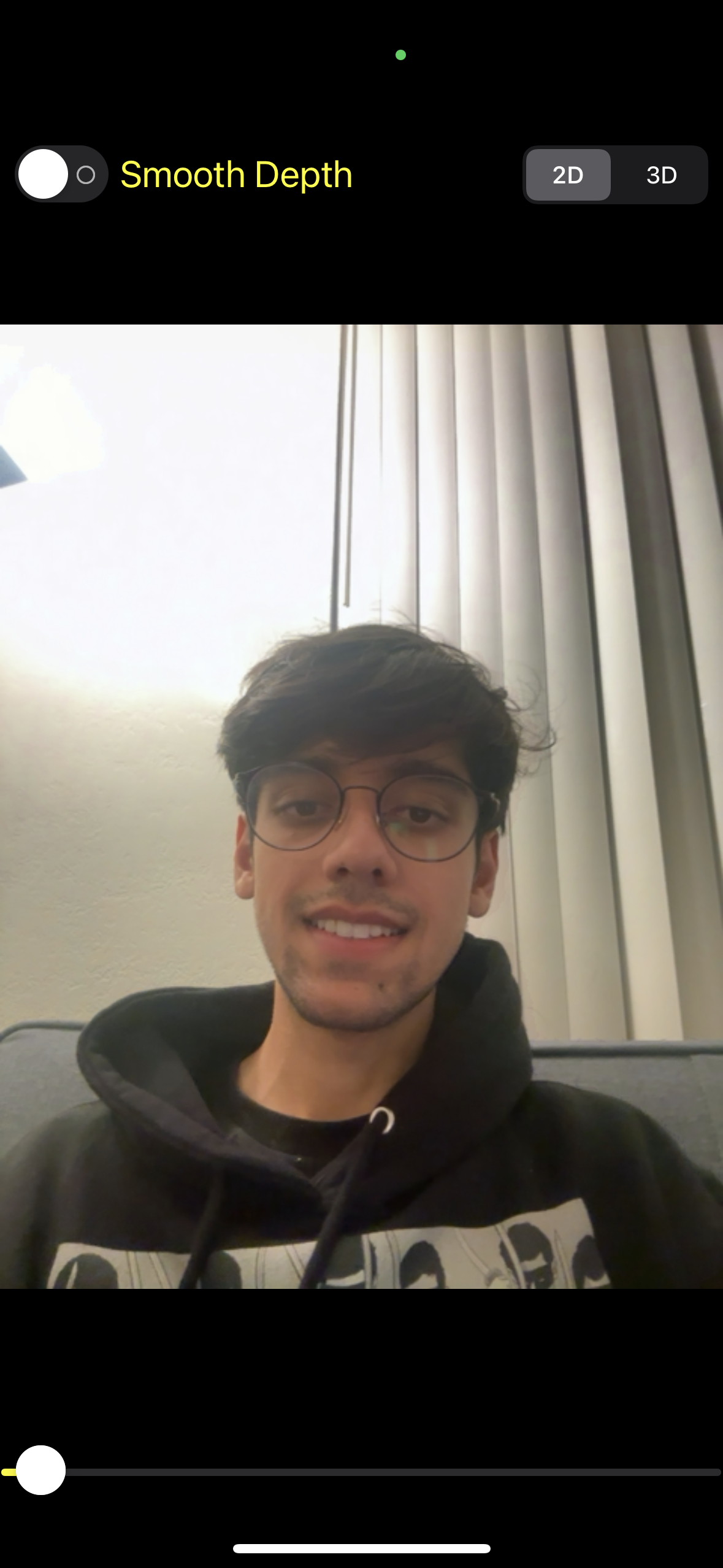}
         \caption{Raw image frame}
         \label{fig:depth1}
     \end{subfigure}
     \hfill
     \begin{subfigure}[t]{0.15\textwidth}
         \centering
         \includegraphics[width=\textwidth]{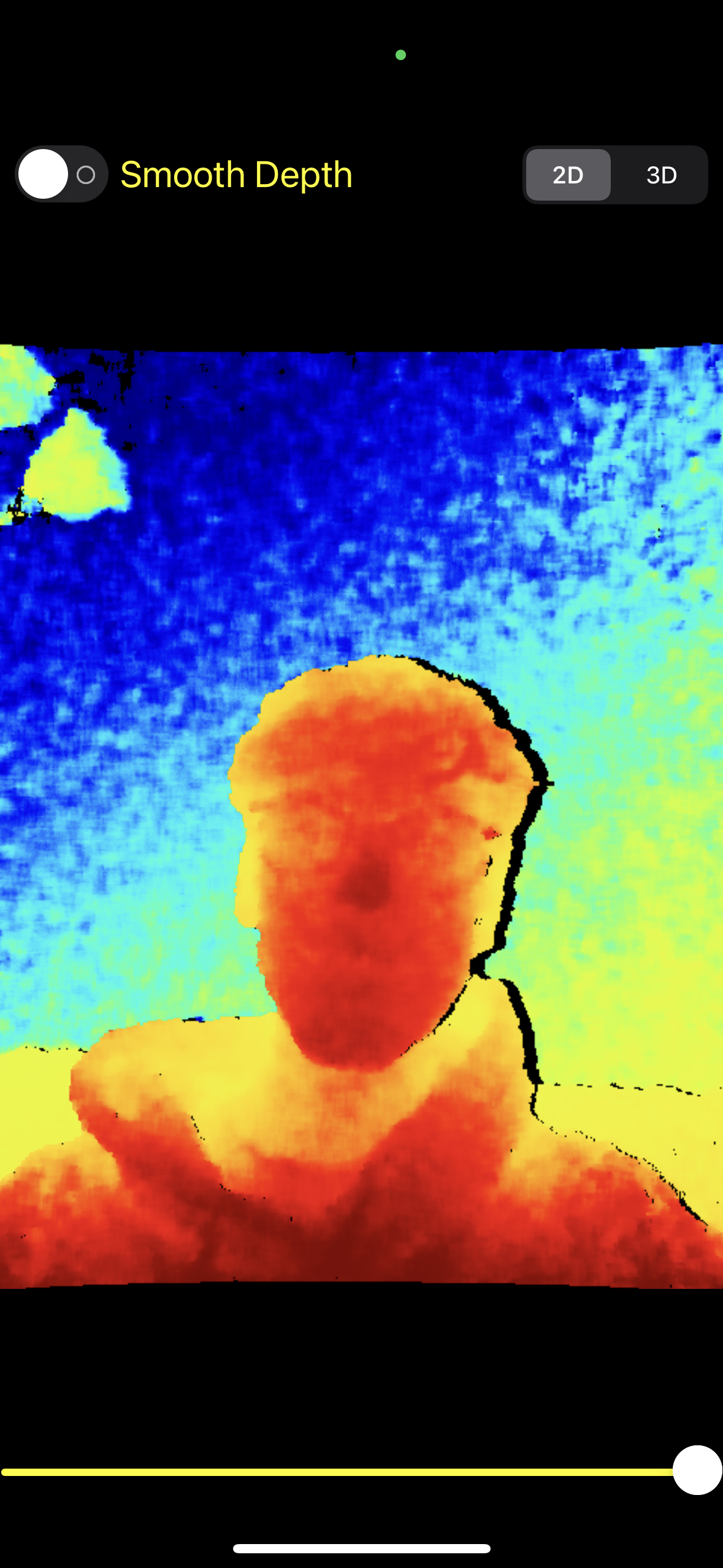}
         \caption{LiDAR + True Depth fused frame upscaled with bi-linear interpolation}
         \label{fig:depth2}
     \end{subfigure}
     \hfill
     \begin{subfigure}[t]{0.15\textwidth}
         \centering
         \includegraphics[width=\textwidth]{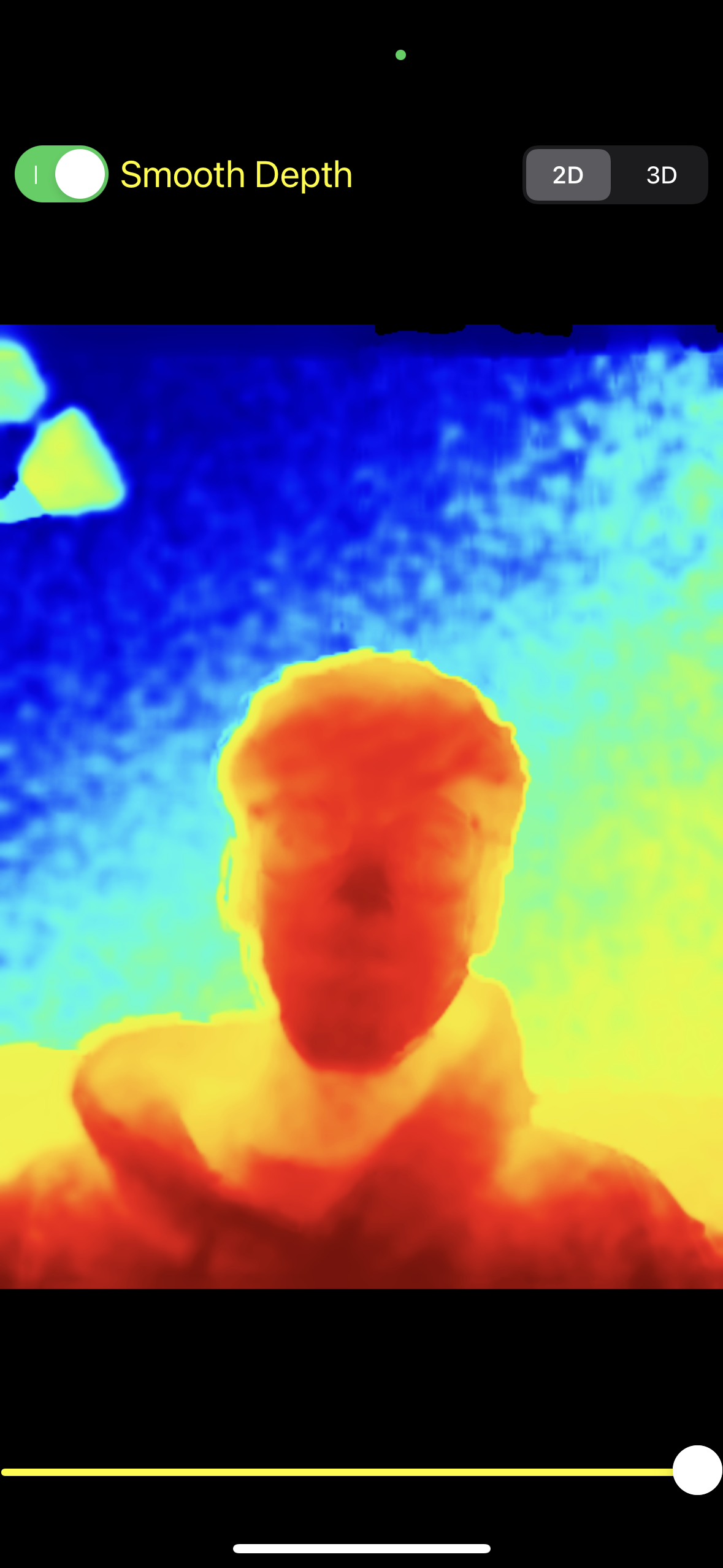}
         \caption{LiDAR + True Depth fused frame upscaled with SRCNN model}
         \label{fig:depth3}
     \end{subfigure}
    \caption{Lidar + TrueDepth data from iPhone 14 Pro generated from reconstruction app programmed in Swift + Metal}
    \label{fig:lidardepth}
\end{figure}

For the second attempt, as shown in Figure \ref{fig:attempt2}, I replace the monocular depth estimation stack with the LiDAR sensor stack, using an iPhone 14 Pro as my test device. This required a rewrite in Swift in order to leverage the native Apple SDKs for GPU and LiDAR programming. Due the low resolution of the LiDAR sensor, a custom trained SRCNN model was employed to upscale the LiDAR depth frame. This model was initially pre-trained on the Set5 super-resolution dataset \cite{dong2015image} and then fine-tuned on a custom dataset of 10,000 face depth images. This dataset was constructed by using the Generated Photos Academic Dataset and feeding it through Marigold with 15 denoising steps for a depth frame. These images were then down scaled by a factor of 4 and anti-aliased for their lower resolution representations. The trained model was deployed via CoreML in order to leverage the Apple Neural Engine on the iPhone \cite{ahremark2022benchmarking}. The results of SRCNN deployed to the LiDAR collection stack is shown in Figure \ref{fig:lidardepth}. Figure \ref{fig:depth2} shows the depth results after a 4x upscale via basic bilinear interpolation. The occluded edges are more dramatic and the overall texture around the edges is more noisy. Figure \ref{fig:depth3} shows the depth results after a forward-inference through the SRCNN model, which shows a more consistent and higher fidelity depth rendering with a 4x upscaler. These screenshots were collected via my Swift application.

After passing the LiDAR data through the model and fusing it with the TrueDepth sensor data via weighted averaging, I used the classical transformation equations from above in order to project the voxels into their world coordinate representations. However, this time, instead of ignoring the extrinsic transformations, I was able to use the IMU sensor on the iPhone to extract the extract camera rotation and translation for the final transformations. I used Metal shaders to render the reconstructed voxel map in real-time via Apple's GPU in my app. 

\begin{figure*}
    \centering
    \includegraphics[width=\textwidth]{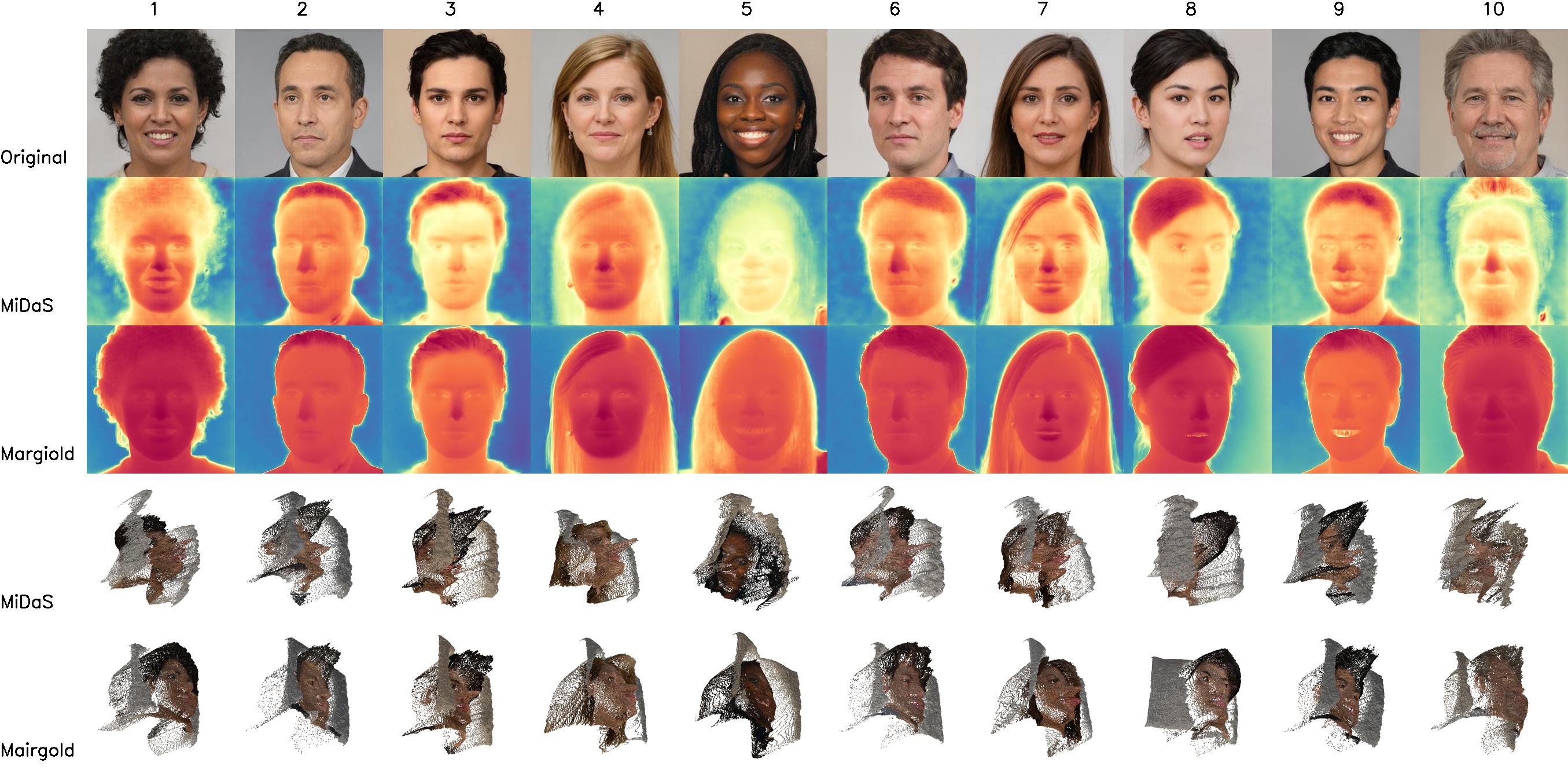}
    \caption{Reconstruction via monocular depth estimation with different modle backbones. In this figure we show MiDaS with low fidelity depth estimations and Marigold as high quality depth estimation.}
    \label{fig:monoculardepth}
\end{figure*}

\subsection{Attempt 3: Template Modeling}

The final attempt, as shown in Figure \ref{fig:attempt3}, was to leverage a deep-learning approach to render a facial reconstruction and apply texture + mesh transformations directly through an end-to-end approach. The intuition behind this approach was that a deep learning method would enable us to learn the general shape of a face and apply internal transformations to render real-time facial reconstructions given a webcam stream. This would potentially allow for more dynamic and responsive interactions in virtual environments, as the system could adjust and adapt to the user's facial expressions and movements in real time. The initial goal was to use INSTA for volumetric avatar rendering given a single RGB frame, however the inference time was too slow for real-time rendering \cite{Zielonka2022InstantVH}. I instead used MICA, which tracks facial landmarks on a face, and then renders a 3D mesh that skews the landmarks from a global mean \cite{Zielonka2022TowardsMR}. I modified this model slightly to use the fused and upscaled LiDAR + TrueDepth sensor data from the prior attempt to aid the model with more accurate metric reconstructions. I retrained MICA on the original dataset, which were a combination of images from LYHM \cite{pears2016automatic}, and FaceWarehouse \cite{cao2013facewarehouse}, but fused a depth dimension generated via Marigold with 10 de-noising steps. 

For real-time rendering, I pass the RGB image frame along with the pixel-wise depth frame through my modified version of MICA. This inference produces a mesh that I rasterize to receive its voxelized representation. Using the depth frame as a heuristic, I project pixels in the background into the rendered scene. Once again using the iPhone's IMU sensor to extrapolate camera rotation and translation to apply final transformation to the rendered voxel grid. As noted in the Results sections, this approach wasn't fully deployed on the iPhone due to some technical issues. Therefore, I was unable to fully render the scene which this approach and was limited to reconstructing the face.

\section{Results}

\subsection{Attempt 1: Monocular Depth Estimation}

\begin{figure}
    \centering
    \includegraphics[width=5cm]{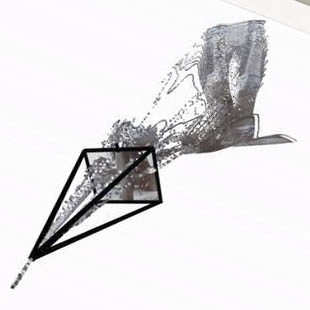}
    \caption{Live reconstruction of a leg and coffee table with MiDAS}
    \label{fig:legs}
\end{figure}

With the monocular depth estimation approach, reconstruction averaged around \textasciitilde 2 FPS with MiDAS and \textasciitilde 0.25 FPS when using Marigold on 10 de-noising steps. The other backbones available on Torch Hub resulted in the same fidelity reconstruction but larger model size, so I defaulted to using the default ViT backbone for my tests. These models were deployed to an iPhone using the PyTorch to CoreML pipeline for quantization and scheduling. An open-source CoreML diffusion model loader application was monkey patched to run inference on these deployed models on an iPhone 14 Pro. For Marigold, decreasing the number of steps introduced quite a few artifacts, and increasing the steps after 15 denoising steps didn't seem to affect the reconstruction quite a lot. Figure \ref{fig:monoculardepth} shows these reconstructions, with the 2nd and 4th row showcasing results of MiDaS reconstructions, and the 3rd and 5th row corresponding to the results of Marigold reconstruction. Figure \ref{fig:legs} shows the result of reconstruction of a pair of legs and a coffee table with the MiDAS model. Unfortunately, offloading data from the open-source app was incredibly buggy therefore the quality of the image is a little low.

From a visual perspective, the main shortcoming I saw with this approach was that the inferred distance from the image plane to the facial features was largely inconsistent. For example, the subject in the 2nd and 4th column of Figure \ref{fig:monoculardepth} are roughly equidistant from the camera. Yet the corresponding reconstructions of these faces are incredibly varied. The 2nd subject has a significantly flatter reconstruction around the cheeks that's not existent in the reconstruction of the 4th subject. Simultaneously we can see that both reconstruction models seemed to have estimated the depth incorrectly for subject in columns 9 and 10. Specifically, the depth model predicted the front cowlicks to extrude outwards which results in a mohawk. From a technical perspective, this makes sense because both MiDaS and Marigold were not trained on a metric accurate datasets. Due to the diversity in the source of the ground truth, all depth values were normalized between 0 to 1, effectively prompting the models to predict relative depth instead of metric accurate depth.

The overall fidelity of reconstructions is not the greatest, and sometimes it's difficult to ascertain whose face is being reconstructed. Increasing the quality of the reconstruction models removes noised by a bit, but sacrifices FPS. We need a way to collect metric accurate sensor data before using our transformation pipeline to project the data. Due to the scope of the project, I wasn't able to allocate enough time + resources to further finetune and quantize the depth estimation models which would have likely helped with increasing the FPS.

\subsection{Attempt 2: LiDAR + TrueDepth}

\begin{figure}
     \centering
     \begin{subfigure}[t]{0.5\textwidth}
         \centering
         \includegraphics[width=\textwidth]{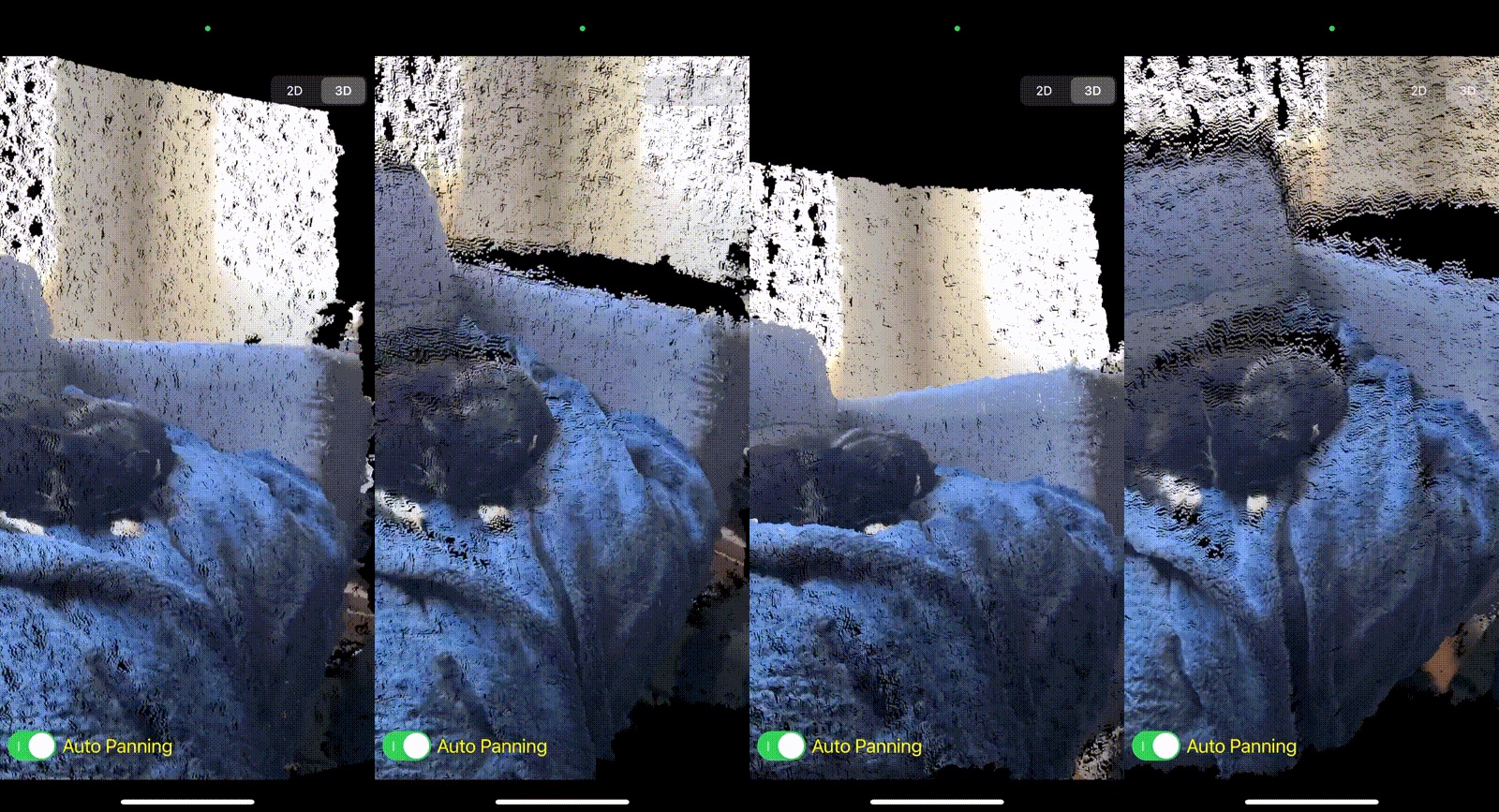}
         \label{fig:lidar1}
     \end{subfigure}
     % \hfill
     \begin{subfigure}[t]{0.5\textwidth}
         \centering
         \includegraphics[width=\textwidth]{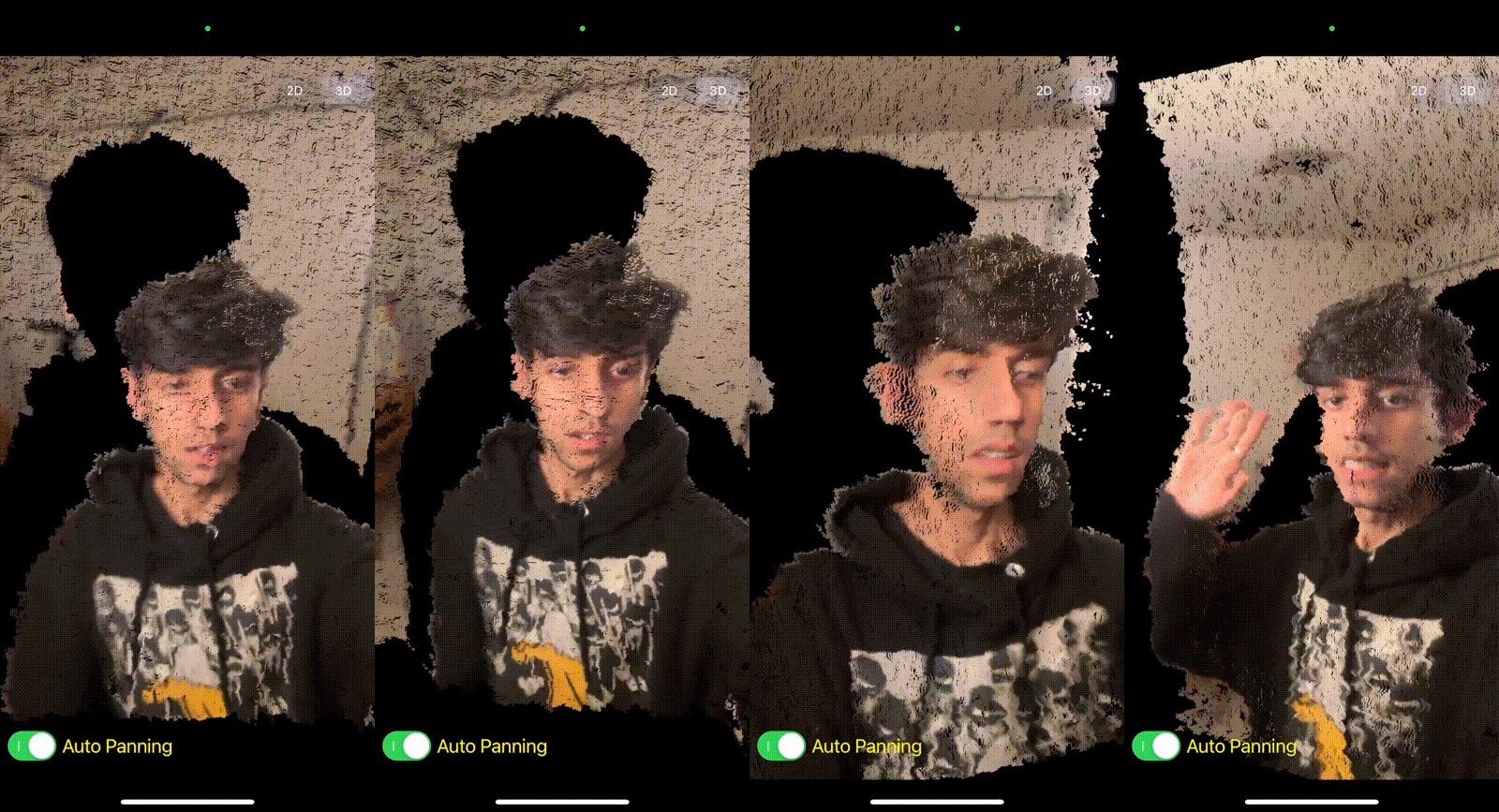}
         \label{fig:lidar2}
     \end{subfigure}
    \caption{Lidar + TrueDepth reconstruction from iPhone 14 Pro}
    \label{fig:lidarrecon}
\end{figure}

Reading the fused LiDAR + TrueDepth data, passing it through the custom upscaled model, and performing the transformations averaged about \textasciitilde 50 frames per second. The main reason behind the throughput being tripled with this approach, was that the majority of the data collection was being done with live sensors instead of any of kind of large-scale neural network inference. I also leveraged the Metal shaders which offloaded a bulk of the parallel computation to the GPU. The unified memory architecture allowed to me perform zero-copy computations which dramatically reduced non-compute latency. The upscaler ML model was run via CoreML so most of the computation was offloaded to the Apple Neural Engine; a forward inference was almost instantaneous. 

Figure \ref{fig:lidarrecon} shows screenshots of the live reconstruction through the custom Swift app. In general, the fidelity of the reconstructions is insanely high. It is quite clear that my face is being reconstructed in Figure \ref{fig:lidarrecon}. From the naked eye, the proportions for the reconstructions also appear to be metric accurate. Even though the upscaler model was finetuned with mostly facial images, it still performs well in reconstructing other subjects like a couch or a cat.

The main shortcoming of this approach, however, was the inability to render any occluded pixels. Naturally, the easiest technical solution to this would be to increase the number of LiDAR and camera sensors to fuse data from multiple viewpoints. However, due to the self-imposed engineering constraint of using one device, simply increasing the number of imaging sensors feels a little bit like cheating.

\subsection{Attempt 3: Template Modeling}

\begin{figure*}
    \centering
    \includegraphics[width=\textwidth]{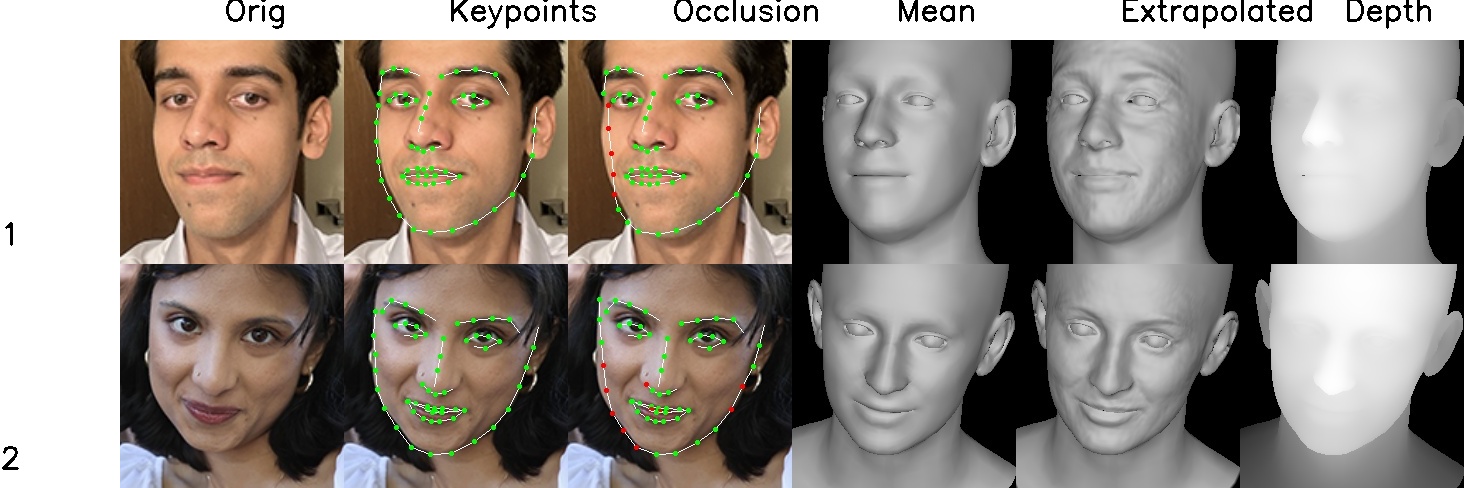}
    \caption{Fused depth + facial landmarks predictions generated by MICA to render 3D mesh of reconstructed face}
    \label{fig:micarecon}
\end{figure*}

\begin{figure}
     \centering
     \begin{subfigure}[t]{0.5\textwidth}
         \centering
         \includegraphics[width=\textwidth]{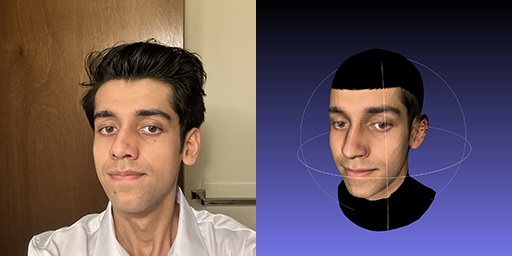}
         \label{fig:ekansh}
     \end{subfigure}
     % \hfill
     \begin{subfigure}[t]{0.5\textwidth}
         \centering
         \includegraphics[width=\textwidth]{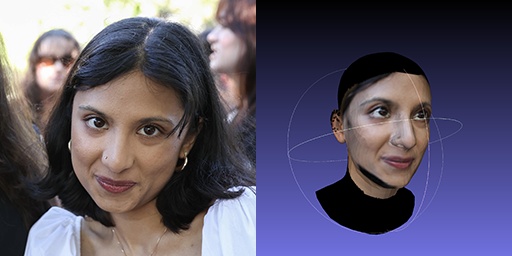}
         \label{fig:lidar}
     \end{subfigure}
    \caption{INSTA end-to-end reconstruction with mesh and texture generated given an image of a face (I can already hear Aleyosha's voice in my head saying "texture is everything" but regardless the reconstructions look incredibly realistic)}
    \label{fig:rendering}
\end{figure}

As an FYI, due to some issues while deploying the CoreML model, this attempt wasn't incredibly reliable as even loading the model would completely crash the iPhone. I had to run these tests on my Nvidia Jetson (8 CPU cores, 32 GB memory, Ampere Nvidia GPU) \cite{karumbunathan2022nvidia} in order to create visualizations for this report, but was still unable to generate GIFs. 

I used saved frames of depth and image data to create the facial reconstructions. Figure \ref{fig:micarecon} visualizes reconstruction via MICA which shows how the facial landmark detection and LiDAR depth estimation come together to render the mesh structure. The figure shows occluded edges through red dots which is important to estimating the face shape for the occluded sections of the images. The corresponding occluded structures on the images are clearly shown in the reconstructions in Figure \ref{fig:rendering}.

The fidelity of these reconstructions is pretty high as shown in Figure \ref{fig:rendering}. The facial features are rendered proportionately; although, I noticed that the MICA more or less ignored rendering the hair. This could be easily solved by fine-tuning MICA with high-fidelity LiDAR scans to measure the depth of individual hair strands. The back of the head is no longer hollow since the model use a texture model as a baseline to extrapolate off of.

I believe that the technical failure of this attempt was more indicative of the limitations of the hardware and software stack rather than the underlying approach itself. Maybe with a bit more of software maturity and/or retooling of the model architecture, I would be able to get more promising results with this hybrid approach over the pure LiDAR-based reconstruction.

\section{Further Study}

Upon achieving decent reconstructions from Attempt 2 with the LiDAR + TrueDepth sensor fusion, it would be interesting to explore basic hardware implementations of projecting holograms into the real-world. Existing research with novel approaches exist \cite{ritter1999hardware} \cite{seo2011cell} \cite{shimobaba2000special} \cite{shimobaba2002special} \cite{spie}, however the ultimate goal should be use a novel approach for multi-dimensional light projection.

\begin{acks}
Special thanks to Aleyosha, Lisa, and Suzie for helping provide the groundwork for this project.
\end{acks}

\bibliographystyle{ACM-Reference-Format}
\bibliography{sample-base}


\end{document}